\documentclass[10pt,a4paper]{article}

\usepackage[a4paper,top=22mm,bottom=23mm,left=25mm,right=25mm,headheight=20pt,headsep=7mm,footskip=11mm]{geometry}
\usepackage[T1]{fontenc}
\usepackage{amsmath}
\usepackage{XCharter}
\usepackage[charter]{newtxmath}
\usepackage[scale=0.90]{sourcesanspro}

\usepackage{pifont}
\usepackage{microtype}
\usepackage{setspace}
\usepackage{graphicx}
\graphicspath{{figures/}}
\usepackage{caption}
\usepackage{subcaption}
\usepackage{booktabs}
\usepackage{array}
\usepackage{tabularx}
\usepackage[skins,breakable]{tcolorbox}
\usepackage{enumitem}

\usepackage[dvipsnames,table]{xcolor}
\definecolor{reportcyan}{HTML}{C36A4A}
\definecolor{reportnavy}{HTML}{214E45}
\definecolor{reportink}{HTML}{24231F}
\definecolor{reportgray}{HTML}{746F66}
\definecolor{reportpanel}{HTML}{F4EFE5}
\definecolor{reportrule}{HTML}{D8CEBC}
\usepackage{tocloft}
\usepackage{fancyhdr}
\usepackage{lastpage}
\usepackage[numbers,sort&compress]{natbib}
\usepackage{placeins}
\usepackage{hyperref}
\hypersetup{
  colorlinks=true,
  linkcolor=reportnavy,
  citecolor=reportnavy,
  urlcolor=reportcyan,
  pdftitle={RecurSciBench: Benchmarking Recursive Self-Improvement in Data-Driven Scientific Research},
  pdfauthor={Wenxi Xu, Chuan Qin, Zihan Wang, Zixing Chen, Le Zhang, Jingbo Zhou, Hao Wang, Yuanchun Zhou, Hengshu Zhu},
  pdfkeywords={Recursive self-improvement, LLM-based agents, data-driven scientific research, benchmarking},
  bookmarksnumbered=true
}

\newcommand{\pipe}{\textsc{RecurSciPipe}}
\newcommand{\bench}{\textsc{RecurSciBench}}
\newcommand{\gevolve}{\textsc{G-evolve}}
\newcommand{\gtest}{\textsc{G-test}}
\def\checkmark{\mbox{\ding{51}}}

\makeatletter
\NewCommandCopy\report@markboth
\DeclareRobustCommand*\markboth[2]{\gdef\report@leftmark{#1}\report@markboth{#1}{#2}}
\newskip\report@skip
\newcommand\report@repeatleftmark{%
  \par
  \ifdim\pagegoal=\maxdimen\else\ifdefined\report@leftmark
    \report@skip\lastskip
    \ifdim\report@skip>\z@\vskip-\report@skip\fi
    \InsertMark{2e-left}{\report@leftmark}%
    \ifdim\report@skip>\z@\vskip\report@skip\fi
  \fi\fi}
\renewcommand\section{\report@repeatleftmark\@startsection{section}{1}{\z@}%
  {-2.15em plus -.3em minus -.2em}{.72em}%
  {\normalfont\Large\bfseries\color{reportnavy}}}
\renewcommand\subsection{\@startsection{subsection}{2}{\z@}%
  {-1.55em plus -.2em minus -.15em}{.45em}%
  {\normalfont\large\bfseries\color{reportink}}}
\renewcommand\subsubsection{\@startsection{subsubsection}{3}{\z@}%
  {-1.2em}{.3em}%
  {\normalfont\normalsize\bfseries\color{reportink}}}
\patchcmd{\NR@sect}{\refstepcounter{#1}}%
  {\LinkTargetOff\refstepcounter{#1}\LinkTargetOn}{}%
  {\PackageError{RecurSciBench}{Could not attach heading destinations}{Check the sectioning definition.}}
\newcommand\report@label@section{{\fontsize{12}{12}\selectfont\sffamily\bfseries\color{reportcyan}\MakeUppercase{\thesection}}\hspace{1em}}
\newcommand\report@label@subsection{{\sffamily\bfseries\color{reportcyan}\thesubsection}\hspace{0.75em}}
\newcommand\report@label@subsubsection{\thesubsubsection\hspace{0.65em}}
\renewcommand\@seccntformat[1]{\MakeLinkTarget{#1}\csname report@label@#1\endcsname}
\newcommand\showarticletitle[1]{#1\report@titleend}
\newcommand\report@titleend{\ifnum\spacefactor>2000 \expandafter\report@eatdot\fi}
\newcommand\report@eatdot{\@ifnextchar.{\@gobble}{}}
\makeatother

\setlist{topsep=0.45em,itemsep=0.2em,parsep=0pt,leftmargin=1.8em}
\setlist[itemize,1]{label=\textcolor{reportcyan}{\raisebox{0.15ex}{\scriptsize\ensuremath{\blacksquare}}}}
\renewcommand{\arraystretch}{1.18}

\renewcommand{\headrulewidth}{0.7pt}
\renewcommand{\footrulewidth}{0pt}
\renewcommand{\headrule}{\hbox to\headwidth{\color{reportrule}\leaders\hrule height \headrulewidth\hfill}}
\renewcommand{\footrule}{\hbox to\headwidth{\color{reportrule}\leaders\hrule height \footrulewidth\hfill}}
\fancypagestyle{plain}{%
  \fancyhf{}
  \fancyhead[L]{\scriptsize\sffamily\bfseries\color{reportnavy}\MakeUppercase{RecurSciBench}}
  \fancyhead[R]{\scriptsize\sffamily\color{reportgray}\nouppercase{\FirstMark{2e-left}}}
  \fancyfoot[L]{\scriptsize\sffamily\color{reportgray}Technical Report}
  \fancyfoot[R]{\scriptsize\sffamily\bfseries\color{reportcyan}\thepage\hspace{0.35em}\color{reportgray}/\hspace{0.35em}\pageref*{LastPage}}
  \renewcommand{\headrulewidth}{0.7pt}
  \renewcommand{\footrulewidth}{0pt}
}

\newif\ifshowtoc
\showtoctrue

\newcommand{\reportlogo}{\includegraphics[height=0.48in]{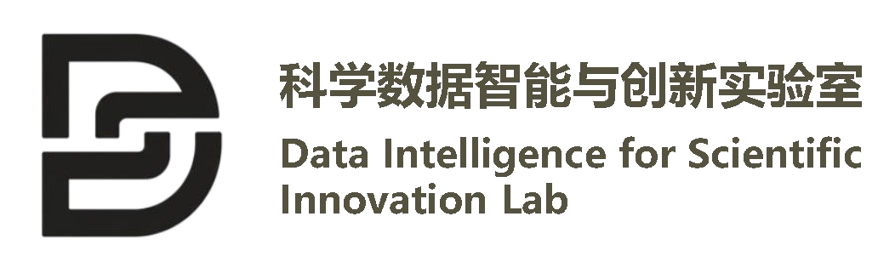}}
\newcommand{\reporttitle}{RecurSciBench: Benchmarking Recursive Self-Improvement in Data-Driven Scientific Research}
\newcommand{\reportauthors}{Wenxi~Xu$^{1}$, Chuan~Qin$^{1,\dagger,*}$, Zihan~Wang$^{1}$, Zixing~Chen$^{1}$, Le~Zhang$^{2}$ \par Jingbo~Zhou$^{2}$, Hao~Wang$^{1}$, Yuanchun~Zhou$^{1}$, Hengshu~Zhu$^{1,*}$}
\newcommand{\reportkeywords}{Recursive self-improvement, LLM-based agents, data-driven scientific research, benchmarking}
\newcommand{\reportemail}[1]{\href{mailto:#1}{\nolinkurl{#1}}}

\newcommand{\custommaketitle}{%
  \thispagestyle{empty}
  \noindent\begin{minipage}[c]{0.66\textwidth}
    \reportlogo
  \end{minipage}%
  \begin{minipage}[c]{0.25\textwidth}
    \raggedleft\small\color{reportgray}\today
  \end{minipage}\par
  \vspace{0.8em}
  \noindent{\color{reportrule}\rule{\textwidth}{0.8pt}}\par
  \vspace{2.3em}
  \noindent{\sffamily\small\bfseries\color{reportcyan}\MakeUppercase{Technical Report}}\par
  \vspace{0.8em}
  \begin{flushleft}
    \sloppy
    {\fontsize{25}{29}\selectfont\bfseries\color{reportnavy}\reporttitle\par}
    \vspace{1.35em}
    {\large\bfseries\color{reportink}\reportauthors\par}
    \vspace{0.5em}
    {\footnotesize\sffamily\color{reportgray}$^{1}$Data Intelligence for Scientific Innovation Lab, Computer Network Information Center, Chinese Academy of Sciences\par}
    {\footnotesize\sffamily\color{reportgray}$^{2}$Baidu Inc.\par}
    {\footnotesize\sffamily\color{reportgray}$^{\dagger}$Project Lead\quad $^{*}$Corresponding Authors\par}
    \vspace{0.32em}
    {\footnotesize\sffamily\color{reportgray}\reportemail{wxxu@cnic.cn}; \reportemail{chuanqin0426@gmail.com}; \reportemail{zhwang@cnic.cn}; \reportemail{zxchen@cnic.cn}; \reportemail{zhangle0202@gmail.com}\par}
    {\footnotesize\sffamily\color{reportgray}\reportemail{zhoujingbo@baidu.com}; \reportemail{cashenry@126.com}; \reportemail{zyc@cnic.cn}; \reportemail{zhuhengshu@gmail.com}\par}
  \end{flushleft}
  \vspace{1.15em}
}

\renewenvironment{abstract}{%
  \begin{tcolorbox}[
    enhanced,breakable,colback=reportpanel,colframe=reportpanel,
    boxrule=0pt,arc=1pt,left=1.15em,right=1.15em,top=0.95em,bottom=0.95em,
    borderline north={2pt}{0pt}{reportcyan}]
  \noindent{\sffamily\bfseries\color{reportcyan}\MakeUppercase{Abstract}}\hspace{0.65em}%
  \color{reportink}\ignorespaces
}{%
  \end{tcolorbox}
  \vspace{0.5em}
}

\begin{document}
\color{reportink}
\custommaketitle

\begin{abstract}
Large language model (LLM)-based agents are advancing the automation of data-driven scientific research. Recent advances in recursive self-improvement (RSI) offer the potential for these agents to continuously improve through experience, but evaluating transfer across scientific tasks requires executable, scientifically verifiable tasks and independent assessment of self-improvement gains. We introduce \bench{}, a scalable benchmark for evaluating RSI in data-driven scientific research. Its \pipe{} framework automatically constructs scientific modeling and data analysis tasks from raw datasets, with execution-based verification and scientific validity checks. \bench{} spans four scientific domains, multiple data modalities, and varied scientific objectives. We evaluate four RSI systems by comparing their initial and evolved states on independent tasks after evolution on source tasks, covering within-type, cross-type, and cross-domain transfer. Experiments show that the benefits of self-improvement vary across agent systems and scientific domains. Positive overall gains can coexist with regressions on individual tasks. Experience can transfer across task types and benefit other domains even without measurable gains in the source domain.
\end{abstract}

{\noindent\small\sffamily\textbf{Keywords:} \reportkeywords\par}
\vspace{0.4em}

\ifshowtoc
  \clearpage
  \thispagestyle{fancy}
  \setcounter{tocdepth}{2}
  \tableofcontents
  \clearpage
\fi

\begin{figure}[t]
\centering
\includegraphics[width=\columnwidth]{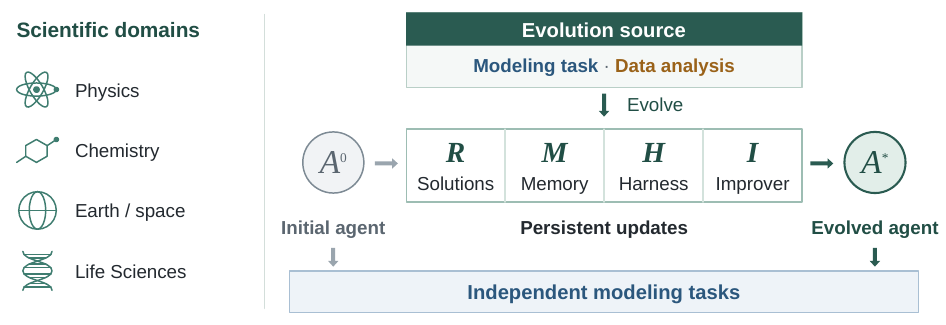}
\caption{RSI evolution and transfer evaluation in \bench{}. Agents acquire experience from scientific analysis and modeling tasks, update persistent components, and are evaluated in their initial and frozen evolved states on independent modeling tasks.}
\label{fig:rsi_evolution}
\end{figure}

\section{Introduction}
\label{sec:introduction}

Data-driven approaches play a central role in scientific research, enabling researchers to uncover patterns and relationships through scientific data analysis and quantitative modeling \citep{fourthparadigm2009,wang2023scientific}. Recent advances in large language model (LLM)-based agents are increasingly automating these research workflows, from data interpretation and method selection to implementation and validation \citep{scienceagentbench2025,ghareeb2026multiagent}. Beyond automating individual research tasks, recursive self-improvement (RSI) enables agents to accumulate experience and iteratively update their own problem-solving mechanisms, including persistent memory, execution harnesses, and improvement strategies \citep{evoscientist2026,hyperagents2026,sia2026}. In data-driven scientific research, such updates offer the potential to transfer knowledge and strategies acquired from previous investigations to new analysis and modeling tasks. However, whether such persistent updates translate into generalizable performance gains on independent scientific tasks remains unclear.

Systematically evaluating recursive self-improvement in data-driven scientific research requires addressing three key challenges. \textbf{(1)~Coverage and extensibility across scientific settings.} Benchmarks for LLMs and agents have progressed from knowledge-based questions to scientific research tasks \citep{gpqa2024,scienceagentbench2025,naturebench2026}. However, scientific research spans diverse domains, data modalities, and modeling assumptions, while many existing benchmarks draw on expert-curated or paper-derived tasks, limiting their extensibility to new scientific datasets and research objectives. \textbf{(2)~Ensuring scientific validity beyond executability.} Executable tasks and automated scoring alone do not guarantee reliable scientific evaluation. Inappropriate modeling assumptions, inconsistent data processing, or unsuitable evaluation criteria can produce misleading results, requiring checks of both execution correctness and scientific validity. \textbf{(3)~Evaluating experience transfer across research granularities.} Although existing RSI benchmarks have explored iterative improvement and cross-task experience reuse in general-purpose agent settings \citep{skillflow2026,skillevolbench2026,evoagentbench2026}, the transfer of experience across different granularities of scientific research remains underexplored. In practice, scientific investigations often involve activities of varying scope, making it challenging to assess whether experience acquired from more focused tasks can improve performance on more comprehensive research problems.

To address these challenges, we introduce \bench{}, a benchmark for evaluating recursive self-improvement in data-driven scientific research. We develop \pipe{}, a four-agent framework that automatically constructs executable scientific data analysis and modeling tasks from raw scientific datasets. Starting with scientific data profiling, \pipe{} jointly designs and builds related analysis and full modeling tasks that share scientific context while retaining distinct objectives and evaluation criteria. The generated tasks undergo execution checks and scientific validity assessments to verify their executability, information isolation, and consistency with scientific requirements. The resulting benchmark spans four scientific domains and multiple data modalities, providing different sources of scientific experience for agent evolution and independent tasks for evaluating its transfer.

Using \bench{}, we evaluate four RSI systems while preserving their native self-improvement and task-solving workflows under fixed backbone model weights. Each system evolves on designated source tasks, after which its frozen evolved state is compared against the corresponding initial state on the same independent tasks (Figure~\ref{fig:rsi_evolution}). We consider three distinct transfer settings: within-type, cross-type, and cross-domain transfer, examining how self-improvement outcomes vary across experience sources and scientific domains. We find that evolution outcomes differ across agent systems and scientific domains, with half of the evolved states showing negative overall gains. The same evolved state can improve performance in one domain while reducing it in another. Positive overall gains can coexist with regressions on a majority of test tasks. Data analysis experience may improve modeling even when evolution on modeling tasks yields a negative gain. Experience from one domain can also benefit tasks in other domains even without measurable gains in the same domain.

Our key contributions are as follows:
\begin{itemize}
    \interlinepenalty=10000
    \item We introduce \bench{}, an extensible RSI benchmark for data-driven scientific research, covering four scientific domains and multiple data modalities. Disjoint evolution and test pools enable frozen-state evaluation of self-improvement transfer.
    \item We develop \pipe{} to automatically construct linked scientific analysis and modeling task families with data, scorers, and task descriptions, verified for executability and scientific validity.
    \item We evaluate four RSI systems across within-type, cross-type, and cross-domain transfer settings. We assess performance gains and regressions and report average token consumption per test task.
\end{itemize}

\section{Related Work}
\label{sec:related_work}

\subsection{Benchmarks for LLMs and Agents in Scientific Research}
\label{sec:scientific_benchmarks}
Benchmarks for LLMs and agents in scientific research evaluate capabilities ranging from scientific knowledge and reasoning to the execution of computational research tasks \citep{gpqa2024,scicode2024,scienceagentbench2025,discoverybench2025}. SciCode and ScienceAgentBench evaluate scientific coding and research tasks derived from expert contributions and published studies \citep{scicode2024,scienceagentbench2025}. Complementary data-science benchmarks, including DSBench, DA-Code, DataSciBench, and MLE-bench, assess agents on data exploration, programming, and machine learning workflows \citep{dsbench2025,dacode2024,zhang-etal-2026-datascibench,mlebench2025}.

Recent approaches have further automated task construction \citep{d3gym2026}. MLE-Smith generates competition-style machine learning engineering tasks, while AutoSDT collects scientific coding tasks from research workflows and synthesizes task instructions and solutions \citep{mlesmith2025,autosdt2025}. These efforts expand the availability of executable tasks, but do not explicitly organize scientifically related analysis and modeling tasks covering multiple scientific domains as distinct sources of research experience. In contrast, \pipe{} jointly constructs such task families from raw scientific data, with independent objectives and evaluation criteria, to support the study of experience transfer across research granularities.

\subsection{Recursive Self-Improvement Agents and Benchmarks}
\label{sec:self_improving_agents}
Recursive self-improvement (RSI) allows LLM-based agents to revise persistent components of their systems through experience and task feedback. Memory-based approaches accumulate research experience for subsequent tasks \citep{evoscientist2026,internagent2026}, while harness-level approaches modify agent implementations, prompts, tools, or execution workflows \citep{dgm2025,sia2026,aflow2025}. HyperAgents further enables modification of the improvement strategy itself \citep{hyperagents2026}, whereas SIA supports both harness and backbone model weight updates \citep{sia2026}. These systems differ in their update scopes and evolution mechanisms, motivating evaluation under a common protocol while preserving their native workflows.

Existing benchmarks assess different aspects of agent self-improvement. RSI-Exam and AI4AI-Bench evaluate iterative research and algorithm optimization, while RSIBench-Data studies data synthesis for improving target models \citep{rsiexam2026,ai4aibench2026,rsibenchdata2026}. SkillFlow and SkillEvolBench examine accumulation and reuse of procedural skills \citep{skillflow2026,skillevolbench2026}, whereas SEAGym and EvoAgentBench investigate the transfer of evolved capabilities to held-out tasks \citep{seagym2026,evoagentbench2026}. Complementing these efforts, \bench{} evaluates how experience accumulated from diverse scientific tasks affects agent performance on new scientific tasks across four scientific domains.

\section{RecurSciPipe Framework}
\label{sec:framework}

\begin{figure*}[t]
\centering

\includegraphics[width=\textwidth]{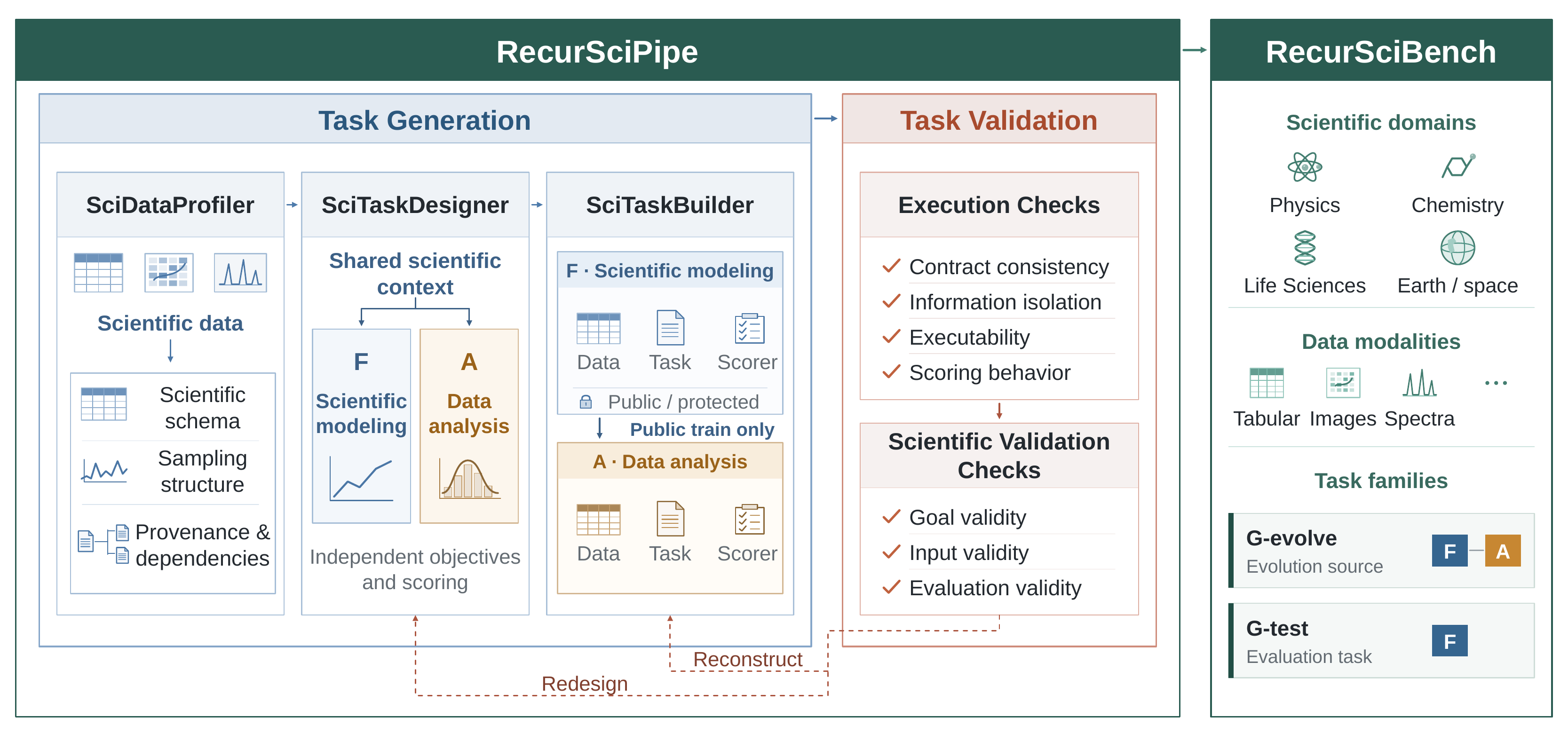}

\caption{Overview of \pipe{} and \bench{}.
\pipe{} generates paired scientific modeling (F) and data analysis (A) tasks, then applies execution and scientific validation checks with feedback for redesign or reconstruction.
\bench{} spans scientific domains and data modalities, with F/A pairs in \gevolve{} for evolution and F tasks in \gtest{} for evaluation.}
\label{fig:generation}
\end{figure*}

Data analysis and scientific modeling are central to data-driven scientific research. Data analysis examines patterns, relationships, and uncertainty in scientific observations, while modeling builds quantitative models using numerical, statistical, or machine learning methods. We introduce \pipe{}, a four-agent framework that constructs executable tasks for RSI evaluation in both activities from raw scientific datasets. As shown in Figure~\ref{fig:generation}, SciDataProfiler, SciTaskDesigner, and SciTaskBuilder profile the data, design tasks, and build executable packages. SciTaskReviewer checks each task's executable behavior and contracts before assessing its scientific objectives, inputs, and evaluation.

\subsection{Unified Generation of Scientific Modeling and Analysis Tasks}
\label{sec:generation}

\noindent\textbf{SciDataProfiler.} For each raw scientific dataset $\mathcal{D}_j$, the agent inspects observations and metadata in their native formats to produce a scientific data profile $P_j$. The profile describes three aspects: the scientific schema of objects, measured quantities, and units; sampling structure across objects, conditions, and time, including repeated observations, missingness, and uncertainty; and provenance and dependencies among sources, observations, and derived quantities. These properties constrain scientific objectives, input availability, and evaluation units in subsequent task design.

\noindent\textbf{SciTaskDesigner.} Given the scientific data profile $P_j$, the agent jointly designs a task family $\mathcal{T}_j=(F_j,A_j)$, comprising a scientific modeling task $F_j$ and a related data analysis task $A_j$. It defines $F_j$'s objective from the scientific context and its targets from measurements, annotations, or documented derivations in $\mathcal{D}_j$. It assesses input availability at inference, data completeness, measurement reliability, and the adequacy of the independent sample size. It then specifies held-out selection criteria and size, accounting for object and temporal dependencies. Evaluation targets reconstruction or interpolation within observed settings, generalization to new objects or observations, or, where supported by the data, out-of-distribution performance under shifts in conditions or input ranges. It designs $A_j$ around a bounded question about measurement semantics, comparability, dependence, or uncertainty. The joint plan specifies each task's permitted inputs, submission contract, and scoring criteria. The tasks share scientific context while retaining distinct objectives and independent evaluation, providing related experience at different granularities.

\noindent\textbf{SciTaskBuilder.} The agent implements the joint plan for $\mathcal{T}_j$ in isolated sessions, constructing $F_j$ before $A_j$ with independent scoring. It prepares public data from the source dataset while preserving its structure, units, and observation identifiers. For $F_j$, it applies the planned split to construct public training data and test inputs while keeping held-out targets private for evaluation. It constructs $A_j$'s inputs and grading evidence exclusively from $F_j$'s public training data. Each package includes a task description, a scorer, and public data.

\subsection{Execution and Scientific Validation Checks}
\label{sec:validation}

\noindent\textbf{SciTaskReviewer.} This agent examines $F_j$ and $A_j$ against the joint plan in independent, concurrent sessions. Each session proceeds to the three scientific checks after completing all four execution checks. Each check records a pass/fail verdict, with a cause and required revision for each failure.

\noindent\textbf{Execution Checks.} The reviewer validates reference solutions for $F_j$ and $A_j$, with $F_j$'s reference score used for normalization (Equation~\ref{eq:normalized_score}). Four rules assess execution and contract correctness:
\begin{enumerate}
\setlength{\itemsep}{0pt}
\setlength{\parsep}{0pt}
\item \textit{Contract Consistency.} The reviewer compares each task package with the joint plan to verify consistency in the provided data, submission requirements, and scoring criteria. It also checks that the public instructions accurately describe the implemented task.
\item \textit{Information Isolation.} The reviewer inspects public materials for leakage of held-out targets or reference answers. It also traces $A_j$'s inputs and grading evidence to verify that they derive only from $F_j$'s public training data.
\item \textit{Executability.} For each of $F_j$ and $A_j$, the reviewer selects a reference solution suited to the task's scientific objective and data. It executes the solution using only permitted inputs and scores the resulting submission to verify that the task can be completed and evaluated as specified.
\item \textit{Scoring Behavior.}
The reviewer evaluates the scorer using test submissions
that vary in answer correctness and format validity. It compares the resulting scores and rejection decisions against independently derived expectations to verify compliance with the published scoring rules.
\end{enumerate}

\noindent\textbf{Scientific Validation Checks.} The reviewer tests the scientific assumptions underlying the constructed tasks:
\begin{enumerate}
\setlength{\itemsep}{0pt}
\setlength{\parsep}{0pt}
\item \textit{Goal Validity.} The reviewer examines the scientific interpretation of the task's targets against source evidence, verifying their derivation and correspondence to observed objects and conditions.
\item \textit{Input Validity.} The reviewer traces inputs to source observations to verify object identity, acquisition timing, and measurement meaning. It examines alignment, unit conversions, and preprocessing dependencies for mismatched observations or use of information unavailable in the declared task setting.
\item \textit{Evaluation Validity.} The reviewer assesses the evaluation population, split, and independent units against the scientific objective. Independent calculations and counterexamples test metrics and tolerances using numerically plausible but scientifically invalid results and valid alternative solutions. Uncertainty calculations are checked against the declared sampling unit and dependence structure.
\end{enumerate}

\noindent\textbf{Repair and Admission.} After all checks are completed, failures requiring design changes are routed first to SciTaskDesigner to revise the plan, after which SciTaskBuilder rebuilds $F_j$ and $A_j$ for review. Failures confined to implementation are routed directly to SciTaskBuilder for rebuilding and review. The task family $\mathcal{T}_j$ is admitted when both tasks pass all checks; otherwise, generation is rejected once the repair-round limit is reached.

\section{RecurSciBench}
\label{sec:benchmark}

\subsection{Task Composition and Splits}
\label{sec:composition}

To support scalable evaluation across diverse scientific settings, we use \pipe{} with GLM-5.3 \citep{zai2026glm53} to construct 58 task families from raw scientific datasets, spanning four scientific domains (Physics, Chemistry, Earth and space sciences, and Life sciences) and diverse data modalities. The benchmark covers varied scientific objectives, such as spectral reconstruction, surface-water forecasting, and identifying speech tasks from patient recordings. We assign 38 task families to the evolution set \gevolve{}, each pairing a scientific data analysis task (A) with a modeling task (F). Another 20 families provide independent modeling tasks for \gtest{}, giving 96 tasks in total. Physics and Earth and space sciences each contribute 10 evolution families, Chemistry and Life sciences 9 each, and every domain contributes 5 modeling tasks to \gtest{}.

\subsection{Evaluation Protocol}
\label{sec:protocol}

\noindent\textbf{RSI Agent Evolution and Evaluation.} To assess whether self-improvement yields transferable capabilities beyond the evolution tasks, we compare each agent's initial and evolved performance on independent tasks. The agent state at iteration $t$ is represented as $A^{t}=(R_t,M_t,H_t,I_t,\theta_t)$, where $R_t$ denotes task-specific solutions and code, $M_t$ memory, $H_t$ the reusable harness of prompts, tools, and workflows, $I_t$ the improvement strategy, and $\theta_t$ the language-model parameters. Starting from $A^{0}$, each system evolves on \gevolve{} by updating a method-specific subset of $R_t$, $M_t$, $H_t$, and $I_t$, while keeping $\theta_t$ fixed. The selected evolved state $A^{*}$ is determined using only evolution-task information and frozen before task evaluation. For each task, independent copies of $A^{0}$ and $A^{*}$ are evaluated on the same task, without carrying task-specific experience across evaluations.

\noindent\textbf{Transfer Settings.} We consider three transfer settings: within-type, cross-type, and cross-domain transfer. Within-type and cross-type transfer use $F$ and $A$ tasks as evolution sources, respectively, with evaluation on the same \gtest{} $F$ tasks. Cross-domain transfer examines generalization when agents evolve on the $F$ tasks of one scientific domain and are evaluated on modeling tasks within and beyond that domain.

\noindent\textbf{Evaluation Metrics.} We report \textit{Mean Task Score} ($S^{*}$) and \textit{Evolution Gain} ($\Delta^{*}$) on \gtest{} modeling tasks. Native metrics differ in scale and optimization direction, so we normalize scores before aggregation. For task $i$ and run $r$, $m_{i,r}^{0}$ and $m_{i,r}^{*}$ denote the scores before and after evolution. Both states are evaluated using up to three valid runs for each task. Let $m_i^{\mathrm{ref}}$ and $o_i$ denote the score of the reference solution and the metric optimum, respectively. The gaps to the optimum are $e_{i,r}^{*}=|m_{i,r}^{*}-o_i|$ and $e_i^{\mathrm{ref}}=|m_i^{\mathrm{ref}}-o_i|>0$. We normalize each run as
\begin{equation}
q_{i,r}^{*}=\frac{e_i^{\mathrm{ref}}}{e_i^{\mathrm{ref}}+e_{i,r}^{*}}.
\label{eq:normalized_score}
\end{equation}
Higher normalized scores are better: the reference scores 0.5 and the optimum 1. For each task, $S_i^{0}$ and $S_i^{*}$ are the mean normalized scores over their $n_i^{0}$ and $n_i^{*}$ valid runs before and after evolution, respectively. We then average the task scores over all $N$ tasks:
\begin{equation}
S_i^{*}=\frac{1}{n_i^{*}}\sum_{r=1}^{n_i^{*}}q_{i,r}^{*},\qquad
S^{*}=\frac{1}{N}\sum_{i=1}^{N}S_i^{*}.
\label{eq:aggregate_score}
\end{equation}
Evolution Gain compares $A^{*}$ with $A^{0}$ by averaging task-level percentage changes:
\begin{equation}
\Delta^{*}=\frac{1}{N}\sum_{i=1}^{N}\frac{S_i^{*}-S_i^{0}}{S_i^{0}}\times100\%,\qquad S_i^{0}>0.
\label{eq:paired_gain}
\end{equation}
We report $S^{0}$, $S^{*}$, and $\Delta^{*}$ on the same \gtest{} tasks, overall and by domain; domain-level metrics use only the tasks in that domain. Positive and negative gains indicate improvement and regression.

\section{Experiments}
\label{sec:experiments}

\subsection{Experimental Setup}
\label{sec:setup}

We evaluate EvoScientist, SIA-H, InternAgent-1.5, and HyperAgents \citep{evoscientist2026,sia2026,internagent2026,hyperagents2026} using GLM-5.3-Flash \citep{zai2026glm53flash} with thinking enabled and high reasoning effort, with model weights fixed. Table~\ref{tab:agent_evolution_objects} summarizes their enabled updates under the state formulation in Section~\ref{sec:protocol}. Agents retain their native evolution procedures and deploy frozen states under three scenarios: within-type, cross-type, and cross-domain transfer. Each system evolves under six source settings in \gevolve{}: all F tasks, all A tasks, and F tasks from each domain, yielding 24 evolved states in total. Each initial and evolved state is evaluated on every \gtest{} task. Each agent is allocated one A800 GPU, 15 logical CPU cores, and 96\,GiB of memory per task, with a six-hour time limit. During task solving, external network access is disabled, and scientific dependencies are available locally. Appendix~\ref{app:agent_details} details the method-specific evolution procedures and configuration.

\begin{table}[t]
\centering
\begingroup
\fontsize{8}{9.2}\selectfont
\setlength{\tabcolsep}{0.7pt}
\renewcommand{\arraystretch}{1.1}
\begin{tabularx}{\columnwidth}{@{}l*{5}{>{\centering\arraybackslash}X}@{}}
\toprule
\textbf{Agent} & \textbf{Solutions} & \textbf{Memory} & \textbf{Harness} & \textbf{Improver} & \textbf{Weights} \\
 & ($R_t$) & ($M_t$) & ($H_t$) & ($I_t$) & ($\theta_t$) \\[1pt]
\midrule
EvoScientist & $\checkmark$ & $\checkmark$ & $\checkmark$ & $\times$ & $\times$ \\
SIA-H & $\checkmark$ & $\times$ & $\checkmark$ & $\times$ & $\times$ \\
HyperAgents & $\checkmark$ & $\checkmark$ & $\checkmark$ & $\checkmark$ & $\times$ \\
InternAgent-1.5 & $\checkmark$ & $\checkmark$ & $\times$ & $\times$ & $\times$ \\
\bottomrule
\end{tabularx}
\endgroup
\caption{Update scopes of the four evaluated RSI systems. $\checkmark$ indicates that a component can be modified during evolution, whereas $\times$ indicates that it remains fixed. All systems use fixed backbone model weights.}
\label{tab:agent_evolution_objects}
\end{table}

\noindent\begin{minipage}{\textwidth}
\centering
\captionsetup{type=table}
\begingroup
\fontsize{7.2}{8.4}\selectfont
\setlength{\tabcolsep}{1pt}
\renewcommand{\arraystretch}{1.12}
\color{reportink}
\arrayrulecolor{reportink}
\begin{tabularx}{\textwidth}{@{}l*{13}{>{\centering\arraybackslash}X}@{}}
\toprule
\textbf{Agent / State} & \multicolumn{5}{c}{\textbf{Mean Task Score}} & \multicolumn{5}{c}{\textbf{Evolution Gain $\Delta^{*}$ (\%)}} & \textbf{W/L} & \multicolumn{2}{c}{\textbf{Tokens (M)}} \\
\cmidrule(lr){2-6}\cmidrule(lr){7-11}\cmidrule(lr){13-14}
& Overall & Phys. & Life & Earth & Chem. & Overall & Phys. & Life & Earth & Chem. & & Input & Output \\
\midrule
\multicolumn{14}{@{}l}{\textbf{EvoScientist}} \\
\quad Initial & 0.5053 & 0.5050 & \underline{0.5001} & 0.5151 & 0.5008 & -- & -- & -- & -- & -- & -- & 4.08 & 0.042 \\
\quad A-evolved & 0.5090 & \underline{0.5266} & 0.4912 & 0.5252 & 0.4928 & \textcolor{green!50!black}{+1.13} & \textcolor{green!50!black}{+5.46} & \textcolor{red}{$-$1.75} & \textcolor{green!50!black}{+2.11} & \textcolor{red}{$-$1.30} & 9/11 & 4.69 & 0.043 \\
\quad F-evolved & 0.5112 & 0.5261 & 0.4911 & 0.5159 & 0.5117 & \textcolor{green!50!black}{+1.42} & \textcolor{green!50!black}{+4.83} & \textcolor{red}{$-$1.82} & \textcolor{green!50!black}{+0.08} & \textcolor{green!50!black}{+2.61} & 9/11 & 4.60 & 0.041 \\
\multicolumn{14}{@{}l}{\textbf{SIA-H}} \\
\quad Initial & \underline{0.5178} & 0.5200 & 0.4987 & \underline{0.5347} & \underline{0.5177} & -- & -- & -- & -- & -- & -- & 2.74 & 0.036 \\
\quad A-evolved & 0.5012 & 0.5135 & 0.4762 & 0.5257 & 0.4920 & \textcolor{red}{$-$2.74} & \textcolor{green!50!black}{+0.42} & \textcolor{red}{$-$4.56} & \textcolor{red}{$-$1.48} & \textcolor{red}{$-$4.72} & 7/12\textsuperscript{*} & 1.96 & 0.046 \\
\quad F-evolved & 0.4995 & 0.4890 & 0.4913 & 0.5249 & 0.4927 & \textcolor{red}{$-$3.34} & \textcolor{red}{$-$5.56} & \textcolor{red}{$-$1.47} & \textcolor{red}{$-$2.15} & \textcolor{red}{$-$4.19} & 8/12 & 6.98 & 0.100 \\
\multicolumn{14}{@{}l}{\textbf{HyperAgents}} \\
\quad Initial & 0.5021 & 0.5080 & 0.4993 & 0.5119 & 0.4892 & -- & -- & -- & -- & -- & -- & 0.90 & 0.024 \\
\quad A-evolved & 0.5037 & 0.5204 & 0.4873 & 0.5262 & 0.4808 & \textcolor{green!50!black}{+0.55} & \textcolor{green!50!black}{+2.89} & \textcolor{red}{$-$2.41} & \textcolor{green!50!black}{+3.67} & \textcolor{red}{$-$1.93} & 9/11 & 0.87 & 0.024 \\
\quad F-evolved & 0.4931 & 0.5012 & 0.4841 & 0.5051 & 0.4822 & \textcolor{red}{$-$1.64} & \textcolor{red}{$-$1.14} & \textcolor{red}{$-$3.03} & \textcolor{red}{$-$0.78} & \textcolor{red}{$-$1.60} & 9/11 & 0.68 & 0.019 \\
\multicolumn{14}{@{}l}{\textbf{InternAgent-1.5}} \\
\quad Initial & 0.4903 & 0.4847 & 0.4772 & 0.5047 & 0.4947 & -- & -- & -- & -- & -- & -- & 12.07 & 0.430 \\
\quad A-evolved & 0.4888 & 0.4993 & 0.4805 & 0.4919 & 0.4834 & \textcolor{red}{$-$0.13} & \textcolor{green!50!black}{+3.09} & \textcolor{green!50!black}{+0.70} & \textcolor{red}{$-$1.94} & \textcolor{red}{$-$2.38} & 10/10 & 14.81 & 0.393 \\
\quad F-evolved & 0.4939 & 0.5026 & 0.4796 & 0.5060 & 0.4873 & \textcolor{green!50!black}{+1.05} & \textcolor{green!50!black}{+4.70} & \textcolor{green!50!black}{+0.52} & \textcolor{green!50!black}{+0.59} & \textcolor{red}{$-$1.62} & 11/9 & 12.64 & 0.433 \\
\bottomrule
\end{tabularx}
\endgroup
\caption{F-evolved states result from evolution on modeling (F) tasks and are evaluated on new modeling tasks (within-type transfer); A-evolved states result from evolution on data analysis (A) tasks and are evaluated on modeling tasks (cross-type transfer). All states are evaluated on the same 20 \gtest{} tasks, five per domain. W/L counts tasks with positive and negative task-level changes. Phys., Life, Earth, and Chem. denote Physics, Life sciences, Earth and space sciences, and Chemistry. Tokens report the average input and output token consumption per test task, in millions (M); output includes reasoning tokens. Underlined values are the highest mean task scores in each column.}
\label{tab:modeling_evolution}

\vspace{12pt}

\centering
\captionsetup{type=table}
\begingroup
\fontsize{7.2}{8.4}\selectfont
\setlength{\tabcolsep}{1pt}
\renewcommand{\arraystretch}{1.12}
\color{reportink}
\arrayrulecolor{reportink}
\begin{tabularx}{\textwidth}{@{}l*{13}{>{\centering\arraybackslash}X}@{}}
\toprule
\textbf{Agent / Source} & \multicolumn{5}{c}{\textbf{Mean Task Score}} & \multicolumn{5}{c}{\textbf{Evolution Gain $\Delta^{*}$ (\%)}} & \textbf{W/L} & \multicolumn{2}{c}{\textbf{Tokens (M)}} \\
\cmidrule(lr){2-6}\cmidrule(lr){7-11}\cmidrule(lr){13-14}
& Overall & Phys. & Life & Earth & Chem. & Overall & Phys. & Life & Earth & Chem. & & Input & Output \\
\midrule
\multicolumn{14}{@{}l}{\textbf{EvoScientist}} \\
\quad Initial & 0.5053 & 0.5050 & 0.5001 & 0.5151 & 0.5008 & -- & -- & -- & -- & -- & -- & 4.08 & 0.042 \\
\quad Physics & 0.5060 & 0.5259 & 0.4889 & 0.5037 & 0.5053 & \textcolor{green!50!black}{+0.39} & \textcolor{green!50!black}{+4.30} & \textcolor{red}{$-$2.22} & \textcolor{red}{$-$1.65} & \textcolor{green!50!black}{+1.13} & 8/12 & 4.15 & 0.044 \\
\quad Life & 0.5110 & 0.5133 & 0.4891 & 0.5344 & 0.5072 & \textcolor{green!50!black}{+1.44} & \textcolor{green!50!black}{+2.20} & \textcolor{red}{$-$2.19} & \textcolor{green!50!black}{+3.89} & \textcolor{green!50!black}{+1.85} & 11/9 & 3.54 & 0.041 \\
\quad Earth & 0.5081 & 0.5344 & 0.4924 & 0.5235 & 0.4819 & \textcolor{green!50!black}{+0.75} & \textcolor{green!50!black}{+6.43} & \textcolor{red}{$-$1.52} & \textcolor{green!50!black}{+1.47} & \textcolor{red}{$-$3.37} & 9/11 & 4.14 & 0.042 \\
\quad Chem. & 0.5159 & 0.5411 & 0.4936 & 0.5288 & 0.5001 & \textcolor{green!50!black}{+2.24} & \textcolor{green!50!black}{+7.28} & \textcolor{red}{$-$1.26} & \textcolor{green!50!black}{+2.73} & \textcolor{green!50!black}{+0.21} & 8/12 & 4.55 & 0.043 \\
\multicolumn{14}{@{}l}{\textbf{SIA-H}} \\
\quad Initial & \underline{0.5178} & 0.5200 & 0.4987 & 0.5347 & 0.5177 & -- & -- & -- & -- & -- & -- & 2.74 & 0.036 \\
\quad Physics & 0.5158 & \underline{0.5424} & 0.4984 & 0.5228 & 0.4961 & \textcolor{red}{$-$0.46} & \textcolor{green!50!black}{+4.26} & \textcolor{green!50!black}{+0.24} & \textcolor{red}{$-$2.49} & \textcolor{red}{$-$3.73} & 9/10\textsuperscript{*} & 5.28 & 0.081 \\
\quad Life & 0.5138 & 0.5138 & 0.4962 & 0.5241 & \underline{0.5210} & \textcolor{red}{$-$0.68} & \textcolor{red}{$-$1.33} & \textcolor{red}{$-$0.50} & \textcolor{red}{$-$2.08} & \textcolor{green!50!black}{+1.17} & 10/10 & 10.48 & 0.106 \\
\quad Earth & 0.5141 & 0.5220 & \underline{0.5064} & 0.5198 & 0.5081 & \textcolor{red}{$-$0.77} & \textcolor{green!50!black}{+0.03} & \textcolor{green!50!black}{+1.53} & \textcolor{red}{$-$2.85} & \textcolor{red}{$-$1.78} & 11/9 & 7.03 & 0.084 \\
\quad Chem. & 0.5089 & 0.5355 & 0.4860 & \underline{0.5417} & 0.4727 & \textcolor{red}{$-$1.65} & \textcolor{green!50!black}{+2.76} & \textcolor{red}{$-$2.51} & \textcolor{green!50!black}{+1.22} & \textcolor{red}{$-$8.07} & 11/9 & 5.33 & 0.149 \\
\multicolumn{14}{@{}l}{\textbf{HyperAgents}} \\
\quad Initial & 0.5021 & 0.5080 & 0.4993 & 0.5119 & 0.4892 & -- & -- & -- & -- & -- & -- & 0.90 & 0.024 \\
\quad Physics & 0.5095 & 0.5334 & 0.4808 & 0.5176 & 0.5063 & \textcolor{green!50!black}{+1.68} & \textcolor{green!50!black}{+5.09} & \textcolor{red}{$-$3.69} & \textcolor{green!50!black}{+1.90} & \textcolor{green!50!black}{+3.43} & 9/11 & 0.94 & 0.024 \\
\quad Life & 0.4887 & 0.4612 & 0.4903 & 0.5305 & 0.4726 & \textcolor{red}{$-$2.48} & \textcolor{red}{$-$8.73} & \textcolor{red}{$-$1.80} & \textcolor{green!50!black}{+4.21} & \textcolor{red}{$-$3.60} & 7/13 & 1.16 & 0.025 \\
\quad Earth & 0.5003 & 0.5037 & 0.4979 & 0.5157 & 0.4841 & \textcolor{red}{$-$0.30} & \textcolor{red}{$-$0.86} & \textcolor{red}{$-$0.31} & \textcolor{green!50!black}{+1.18} & \textcolor{red}{$-$1.21} & 8/12 & 1.12 & 0.026 \\
\quad Chem. & 0.4927 & 0.4897 & 0.4937 & 0.5004 & 0.4871 & \textcolor{red}{$-$1.61} & \textcolor{red}{$-$2.69} & \textcolor{red}{$-$1.13} & \textcolor{red}{$-$1.94} & \textcolor{red}{$-$0.66} & 8/12 & 0.96 & 0.024 \\
\multicolumn{14}{@{}l}{\textbf{InternAgent-1.5}} \\
\quad Initial & 0.4903 & 0.4847 & 0.4772 & 0.5047 & 0.4947 & -- & -- & -- & -- & -- & -- & 12.07 & 0.430 \\
\quad Physics & 0.4935 & 0.5114 & 0.4798 & 0.4953 & 0.4874 & \textcolor{green!50!black}{+1.15} & \textcolor{green!50!black}{+6.61} & \textcolor{green!50!black}{+0.56} & \textcolor{red}{$-$1.52} & \textcolor{red}{$-$1.04} & 11/9 & 12.72 & 0.398 \\
\quad Life & 0.5001 & 0.5221 & 0.4757 & 0.4978 & 0.5048 & \textcolor{green!50!black}{+2.62} & \textcolor{green!50!black}{+9.49} & \textcolor{red}{$-$0.37} & \textcolor{red}{$-$0.99} & \textcolor{green!50!black}{+2.35} & 11/9 & 12.14 & 0.433 \\
\quad Earth & 0.4870 & 0.5000 & 0.4827 & 0.5090 & 0.4561 & \textcolor{red}{$-$0.37} & \textcolor{green!50!black}{+3.71} & \textcolor{green!50!black}{+1.20} & \textcolor{green!50!black}{+1.21} & \textcolor{red}{$-$7.62} & 10/10 & 12.89 & 0.420 \\
\quad Chem. & 0.4943 & 0.5072 & 0.4872 & 0.4794 & 0.5032 & \textcolor{green!50!black}{+1.18} & \textcolor{green!50!black}{+5.39} & \textcolor{green!50!black}{+2.10} & \textcolor{red}{$-$4.60} & \textcolor{green!50!black}{+1.82} & 12/8 & 12.43 & 0.398 \\
\bottomrule
\end{tabularx}
\endgroup
\caption{Each system evolves separately on the F source tasks of one domain and is evaluated on all 20 \gtest{} tasks. Underlined values are the highest mean task scores in each column.}
\label{tab:domain_transfer}
\end{minipage}%
\begingroup
\renewcommand{\thefootnote}{\fnsymbol{footnote}}
\footnotetext[1]{For each starred state, one task without a valid run is excluded.}
\endgroup
\par

\subsection{Overall Performance}
\label{sec:overall_performance}

Tables~\ref{tab:modeling_evolution} and~\ref{tab:domain_transfer} compare the initial and evolved states of each agent system. Overall gains across the 20 \gtest{} tasks from all four domains are small and inconsistent: across the 24 evolved states, $\Delta^{*}$ ranges from $-3.34\%$ to $+2.62\%$, and half of the states have negative gains. Transfer also differs across systems. EvoScientist improves overall in all six evolved states ($+0.39\%$ to $+2.24\%$), whereas SIA-H declines in all six ($-3.34\%$ to $-0.46\%$). InternAgent-1.5 has positive overall gains in four states and HyperAgents in two. At the task level, every evolved state regresses on at least 8 of the 20 test tasks. Some states achieve positive overall gains while regressing on 12 tasks, showing that larger relative gains on a subset of tasks can outweigh losses on the majority. Self-improvement therefore produces uneven transfer across tasks, with the direction of its overall effect depending on the agent system.

\subsection[Within-Type Transfer]{Within-Type Transfer}
\label{sec:modeling_transfer}

In within-type transfer, each agent system evolves on the 38 modeling (F) tasks in \gevolve{} and is evaluated on the 20 modeling tasks in \gtest{} (Table~\ref{tab:modeling_evolution}). Evolution on the same 38 source tasks produces both positive and negative overall gains: EvoScientist ($+1.42\%$) and InternAgent-1.5 ($+1.05\%$) improve, whereas HyperAgents ($-1.64\%$) and SIA-H ($-3.34\%$) decline. Domain-level gains span a wider range, from $-5.56\%$ to $+4.83\%$. Within a system, gains can have opposite signs: EvoScientist improves on Physics by $4.83\%$ but declines on Life sciences by $1.82\%$, while InternAgent-1.5 improves on Physics by $4.70\%$ but declines on Chemistry by $1.62\%$. Both systems improve in three domains, whereas HyperAgents and SIA-H decline in all four. These results show that the direction and magnitude of within-type transfer depend on both the agent system and the scientific domain, with overall gains masking declines in individual domains.

\textbf{Case study.} HyperAgents shows the clearest within-type decline in Life sciences, where its mean task score falls from $0.4993\pm0.0030$ to $0.4841\pm0.0034$ (Table~\ref{tab:test_variation}). During evolution on the F tasks, it stored in memory ($M$) a rule for nearly tied candidates: when several candidates score within one standard error of the best cross-validation score, submit the simplest. The rule was written after cross-validation had overstated the accuracy of selected models on the source tasks, and it applies to any test task on which validation leaves candidates nearly tied. One such task asks for half-hourly CO$_2$ flux in a withheld year (2018) given concurrent meteorology, with three earlier years (2015--2017) for training. All three F-evolved runs validate candidates by leave-one-year-out cross-validation over these years. With three folds, the standard error of each validation score is large, and a climatology, the historical mean for each month and half-hour of day, scores within one standard error of the best candidate or better. The rule therefore applies: two runs submit the climatology, which ignores meteorology, and one averages it with a weather-conditional model. They score 0.468--0.489, below the scores of all three initial runs, which fit gradient-boosted trees to the weather variables. A safeguard against overfitting on the source tasks thus selects an estimator that ignores the covariates on which the new task is conditioned.

\subsection[Cross-Type Transfer]{Cross-Type Transfer}
\label{sec:analysis_transfer}

In cross-type transfer, each agent system evolves on the 38 scientific data analysis (A) tasks in \gevolve{} and is evaluated on the 20 modeling (F) tasks in \gtest{} (Table~\ref{tab:modeling_evolution}). Overall gains across the 20 test tasks are positive for EvoScientist ($+1.13\%$) and HyperAgents ($+0.55\%$), and negative for SIA-H ($-2.74\%$) and InternAgent-1.5 ($-0.13\%$). For HyperAgents, the overall gain changes from $-1.64\%$ under within-type transfer to $+0.55\%$ under cross-type transfer. At the domain level, all four systems gain on Physics under cross-type transfer ($+0.42\%$ to $+5.46\%$), including HyperAgents and SIA-H, which regress on Physics under within-type transfer. These results show that experience acquired from scientific data analysis transfers to scientific modeling and may produce positive gains even when evolution on modeling tasks yields a negative gain.

\textbf{Case study.} Rules learned on A tasks transfer to modeling as working procedures, and whether they help depends on what limits a submission's score. The A tasks grade contract-defined numerical outputs, and EvoScientist's A-evolved harness ($H$) contains a skill that requires an independent verifier to recompute numerical outputs before submission; the A-evolved agent writes such a verifier in 38 of 60 test runs, compared with 2 of 60 for the initial agent. The rule helps when an implementation defect limits the score. On a creep crack growth task that predicts potential-drop ratios, one run's verifier reveals that a bug has made the predictions constant within each specimen; after the fix, the run scores 0.522, above the scores of all three initial runs. The rule does not help when the score depends on the choice of model. On a task that reconstructs masked bands of infrared spectra, all three A-evolved runs verify their predictions before submission, yet score 0.414--0.437, below the scores of all three initial runs: the verifier confirms that the predictions are computed as intended, not that the chosen model is accurate. Over the 20 test tasks, the mean task score stays within test variation ($0.5053\pm0.0058$ before and $0.5090\pm0.0035$ after evolution; Table~\ref{tab:test_variation}).

\subsection[Cross-Domain Transfer]{Cross-Domain Transfer}
\label{sec:domain_transfer}

In cross-domain transfer, each agent system evolves on the modeling (F) tasks of one scientific domain and is evaluated on all 20 \gtest{} tasks across the four domains (Table~\ref{tab:domain_transfer}). Transfer patterns differ markedly across agent systems. EvoScientist achieves positive overall gains across the 20 test tasks for all four source domains ($+0.39\%$ to $+2.24\%$), whereas SIA-H has negative overall gains in all four cases ($-1.65\%$ to $-0.46\%$). At the domain level, EvoScientist and InternAgent-1.5 improve on Physics after evolution on each of the four source domains, with gains of $+2.20\%$ to $+7.28\%$ and $+3.71\%$ to $+9.49\%$, respectively. SIA-H also improves on Physics for three source domains despite its negative overall gains. For HyperAgents, the gain on Physics is positive after same-domain evolution ($+5.09\%$) but negative after evolution on each of the other three domains ($-8.73\%$ to $-0.86\%$). For EvoScientist, evolution on Chemistry tasks yields a gain of $+0.21\%$ on Chemistry and $+7.28\%$ on Physics. These results show that experience can benefit tasks in other domains even when it does not measurably improve performance in the same domain, with transfer outcomes varying across agent systems.

\textbf{Case study.} Cross-domain gains arise when a domain-agnostic rule matches the structure of a task in another domain. After evolution on Chemistry tasks, EvoScientist's Physics score rises from $0.5050\pm0.0125$ to $0.5411\pm0.0111$, while its Chemistry score is unchanged ($0.5008\pm0.0086$ and $0.5001\pm0.0114$; Table~\ref{tab:test_variation}). The gain comes from two Physics tasks that extrapolate beyond the observed range: emission spectra of a doped-glass series at higher coordinate values, and the later force response of laminate tests. A skill learned on chemistry curve tasks requires validation splits to mimic the scored evaluation, and five of the six evolved test runs on these tasks validate by forward extrapolation. All three evolved runs on the laminate task score 0.691--0.751, above the scores of all three initial runs. On the spectral task, validation over the full forecast horizon favors quadratic over linear growth, and two evolved runs that select it score 0.500 and 0.517, above the scores of all three initial runs. Conversely, after evolution on Physics tasks, HyperAgents keeps in memory the lesson that simple smooth models outperform complex ones on small scientific data; on a Life sciences task that estimates leaf chlorophyll from multispectral imagery, all three runs average a ridge regression into their tree ensembles and score below the scores of all three initial runs, and its Life sciences score falls from $0.4993\pm0.0030$ to $0.4808\pm0.0019$.

\subsection{Discussion}
\label{sec:discussion}

Across the three transfer settings, accumulated experience yields uneven benefits on new tasks. Positive overall gains can coexist with regressions on individual tasks. Experience acquired from a different task type may yield positive gains, and experience from one domain can benefit tasks in other domains even when it does not measurably improve performance in the same domain. SIA-H's negative overall gains under all six source settings suggest poor generalization of its learned experience. Assessing the benefit of evolution therefore requires considering the agent system, the source of experience, and the evaluation tasks together.

One explanation for uneven transfer is that source-task optimization makes particular choices persist beyond the conditions in which they are useful. For example, HyperAgents learns on source tasks to ship the simplest of nearly tied candidates, and on a new task this preference removes the covariates on which the task conditions (Section~\ref{sec:modeling_transfer}). Agents should therefore be designed to accumulate experience while retaining the flexibility to choose and adapt solutions for new tasks. Their evolved states should also be evaluated on diverse tasks to assess how well the acquired experience generalizes.

\section{Online Platform}
\label{sec:platform}

\begin{figure}[!t]
\centering
\begin{subfigure}{\textwidth}
  \centering
  \includegraphics[width=\linewidth]{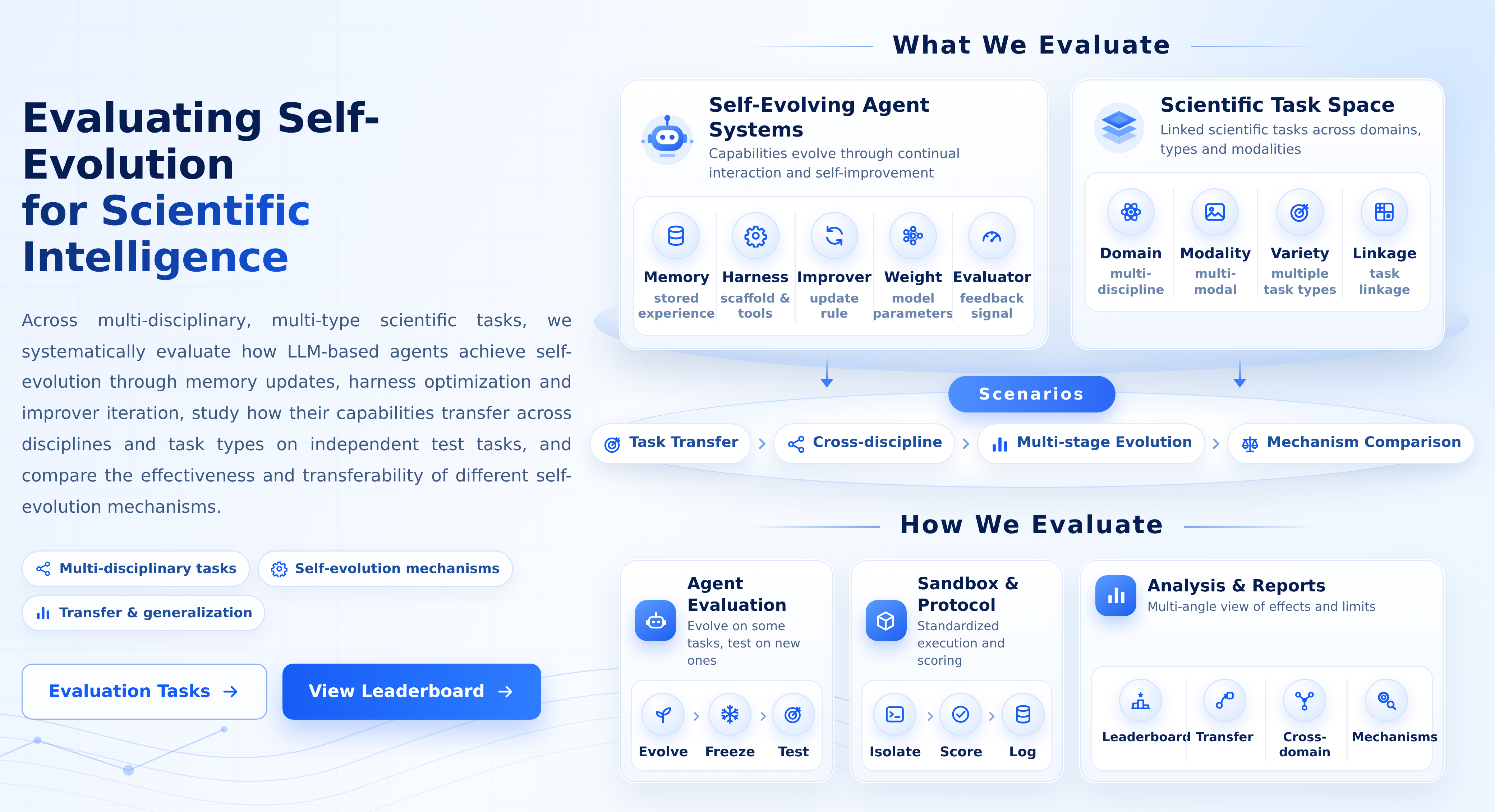}
  \caption{Evaluation goal}
  \label{fig:platform_goal}
\end{subfigure}\par\medskip
\begin{subfigure}[c]{0.355\textwidth}
  \centering
  \includegraphics[width=\linewidth]{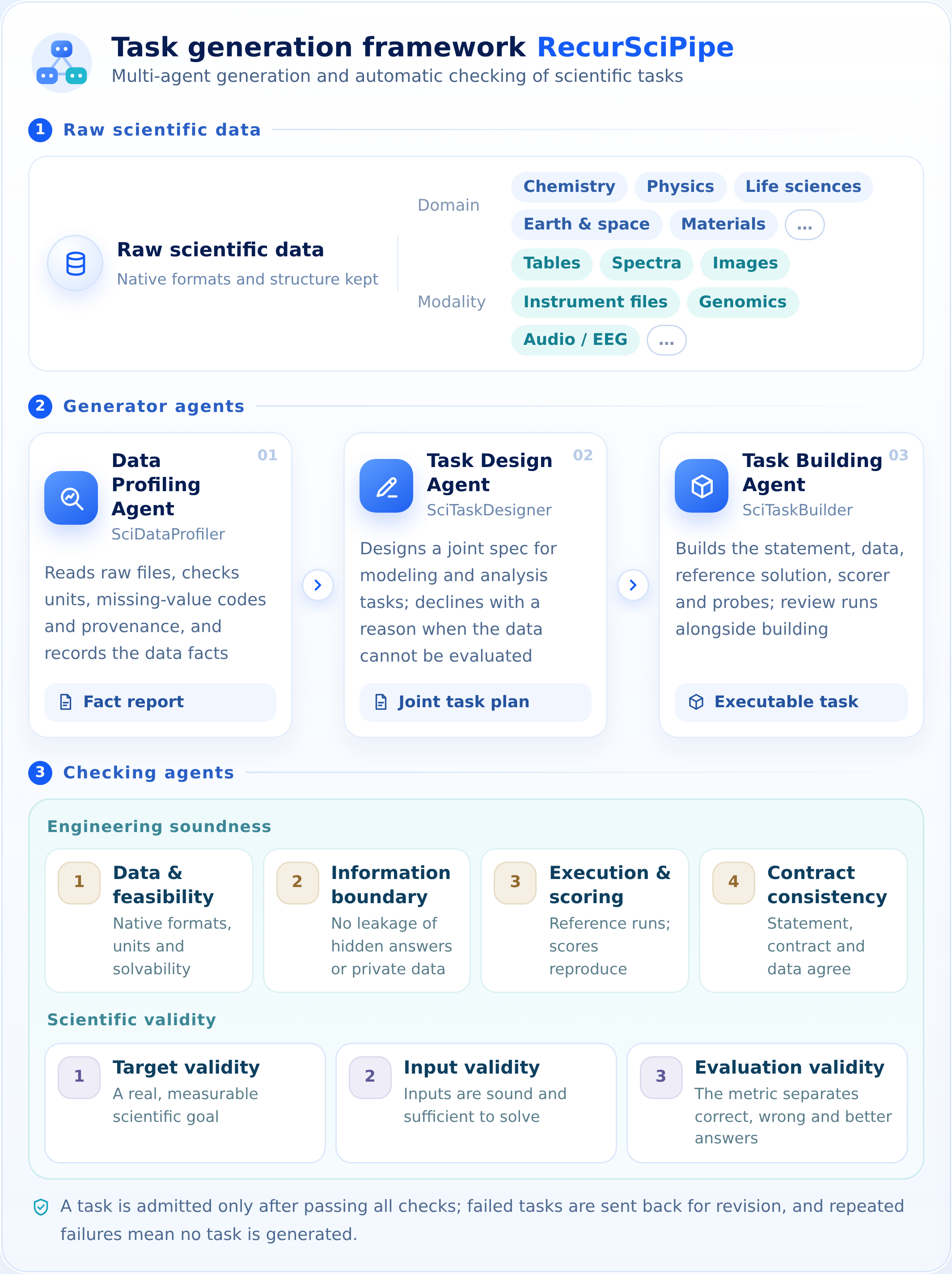}
  \caption{Task generation framework}
  \label{fig:platform_framework}
\end{subfigure}\hfill
\begin{minipage}[c]{0.615\textwidth}
  \begin{subfigure}{\linewidth}
    \centering
    \includegraphics[width=\linewidth]{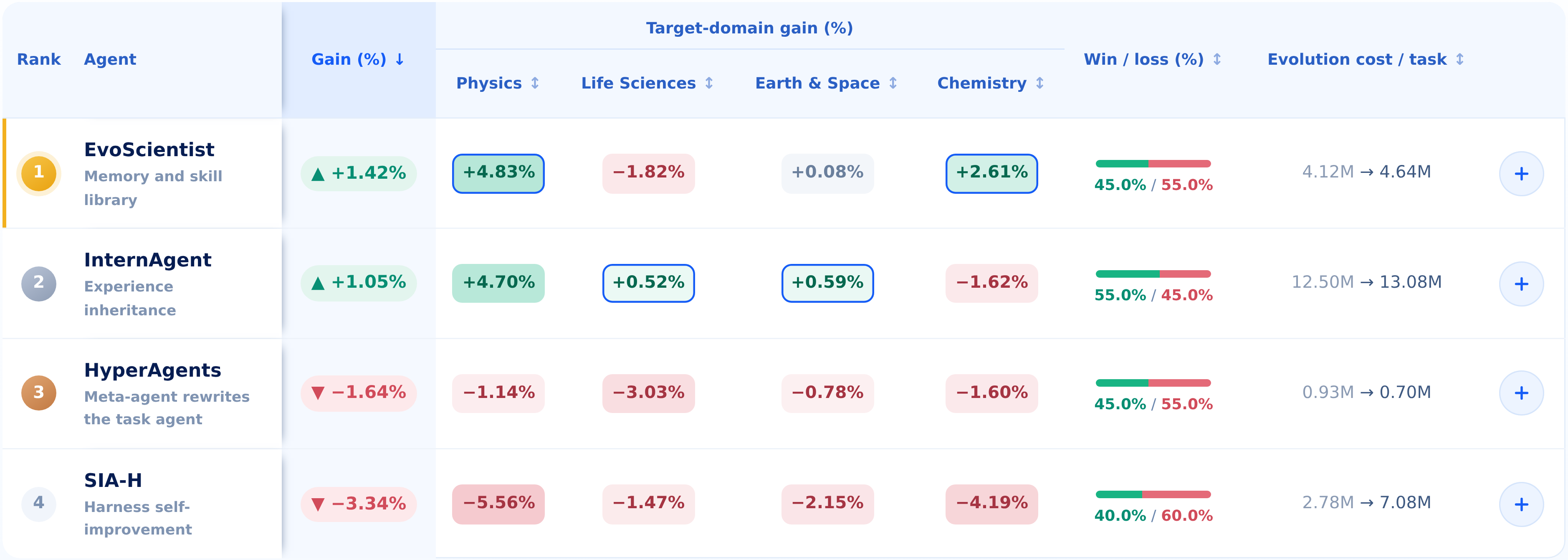}
    \caption{Self-evolution leaderboard}
    \label{fig:platform_leaderboard}
  \end{subfigure}\par\medskip
  \begin{subfigure}{\linewidth}
    \centering
    \includegraphics[width=\linewidth]{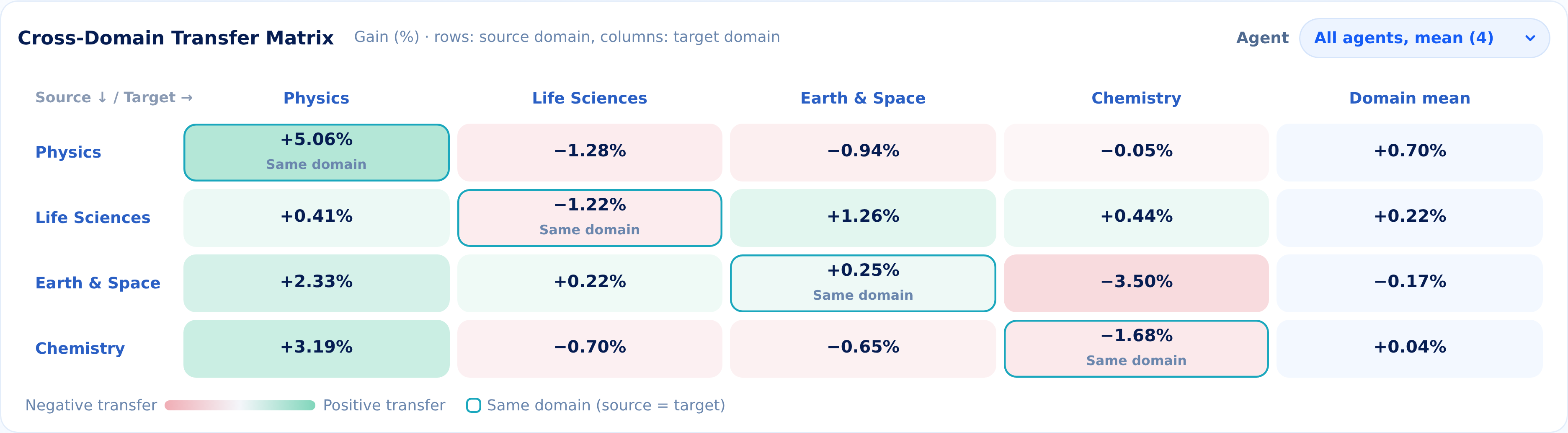}
    \caption{Cross-domain transfer matrix}
    \label{fig:platform_matrix}
  \end{subfigure}
\end{minipage}
\caption{The \bench{} platform. The leaderboard in (c) shows within-type transfer on modeling tasks, and the matrix in (d) averages gains over the four agent systems, with source domains as rows and target domains as columns.}
\label{fig:platform}
\end{figure}

We release \bench{} and its results on our online platform at \url{https://www.scihorizon.cn/SelfEvolution} (Figure~\ref{fig:platform}). The overall goal of the platform is to evaluate self-evolution for scientific intelligence (Figure~\ref{fig:platform_goal}): it covers agent systems that evolve their memory, harness, improvement strategy, and model weights under evaluator feedback, a space of linked scientific tasks across domains, modalities, and task types, and scenarios that include task transfer, cross-discipline generalization, multi-stage evolution, and mechanism comparison.

The platform describes the \pipe{} framework, including its generator and checking agents, and shows example tasks from \bench{}. Its self-evolution leaderboard reports the results of Section~\ref{sec:experiments} in three tracks: within-type transfer on modeling tasks, cross-type transfer from data analysis to modeling, and cross-domain transfer. Each track reports overall and domain-level gains, win/loss rates, and token consumption per task, and the cross-domain track adds a source-by-target transfer matrix for each agent system and for their mean.

\FloatBarrier
\section{Conclusion}
\label{sec:conclusion}

We introduce \bench{} to evaluate how experience acquired from scientific tasks affects performance on new tasks. Its \pipe{} framework constructs paired data analysis and modeling tasks from raw datasets, with execution checks and scientific validation. Using tasks from four scientific domains, we compare the initial and evolved states of four RSI systems under within-type, cross-type, and cross-domain transfer. Evolution yields uneven benefits across systems and tasks: half of the evolved states have negative overall gains, and positive overall gains can coexist with regressions on individual tasks. Data analysis experience may also improve modeling, while experience from one domain can benefit another without measurably improving performance in its source domain. Because the same experience can help one task and hinder another, agents should retain the conditions under which it is useful and adapt its use to the requirements of new tasks.

\section*{Limitations}
\addcontentsline{toc}{section}{Limitations}
\label{sec:limitations}

Our evaluation covers 20 independent modeling tasks across four scientific domains and four RSI systems, all using GLM-5.3-Flash as the backbone. The task and system counts limit coverage of scientific problems and agent designs, and the effect of backbone choice remains untested. Evaluating more tasks, agent systems, and backbone models would clarify how broadly the observed transfer patterns hold.

\clearpage
\FloatBarrier
\appendix
\setcounter{table}{0}
\setcounter{figure}{0}
\renewcommand{\thetable}{S\arabic{table}}
\renewcommand{\thefigure}{S\arabic{figure}}
\renewcommand{\theHtable}{appendix.\arabic{table}}
\renewcommand{\theHfigure}{appendix.\arabic{figure}}
\section{Evolution Procedure and Configuration of RSI Agent Systems}
\label{app:agent_details}

Following Section~\ref{sec:protocol}, $R$ denotes task solutions and code, $M$ stored experience, $H$ the execution harness comprising prompts, tools, and reusable workflows, and $I$ the improvement strategy that generates agent updates. The enabled updates differ across systems, while language-model parameters $\theta$ remain fixed. After source evolution, a selected persistent state is frozen for independent task evaluation.

\noindent\textbf{EvoScientist.} EvoScientist \citep{evoscientist2026} evolves research memory ($M$) and reusable skills ($H$). Its research agent solves source tasks using retained memories and skills, while memory workers consolidate observations and experience. AutoSkills\footnote{AutoSkills is part of the EvoScientist software release (\url{https://github.com/EvoScientist/EvoScientist}).} can synthesize and maintain skills from this research history. The retained memory and skills pass to subsequent source tasks, and the final complete state is selected for evaluation.

We use one learning workflow per source task and an AutoSkills graph limit of 400 steps. The adapter supplies public scientific tasks to the research workspace and invokes AutoSkills after each source task, with automatic installation of validated skill proposals. Its native workflow has no channel for ground-truth scores, so the agent receives no score computed on the held-out targets of source tasks in \gevolve{}; the held-out targets of \gtest{} tasks and their scores are likewise invisible to it.

\noindent\textbf{SIA-H.} SIA-H denotes SIA \citep{sia2026} with harness updates only. It evolves the Target agent's execution harness ($H$). Meta constructs a Target from SIA's reference agent template, and Feedback revises it using execution traces, evaluation results, and the recorded evolution history. These revisions can change the Target's code, prompts, tools, and workflow, while Meta and Feedback remain fixed.

We use the harness mode (\texttt{-{}-focus harness}) with two Target generations per source task and a 120-turn limit for Meta and Feedback. The adapter connects the evaluation and feedback loop to scientific-task grading and carries the reusable Target package across source tasks. As in SIA's native design, each Target generation on a source task in \gevolve{} is scored on the task's held-out targets, and Feedback reads the resulting metric before writing the next generation, so each source task yields two scores. The last complete Target package is frozen and executed directly on each \gtest{} task, whose held-out targets and scores remain invisible to it.

\noindent\textbf{InternAgent-1.5.} InternAgent-1.5 \citep{internagent2026} generates research ideas, implements and evaluates selected experiments, and extracts experience from their outcomes. It accumulates experimental records in memory ($M$), which TaskMemory retrieves during idea generation on subsequent source tasks. The memory after the last completed source task is frozen for evaluation.

We use one discovery round per source task, generate six ideas, select two, and allow up to two workflow iterations. PromptEvolver is disabled. The adapter binds the research workflow to the public scientific objective and submission contract. As in the native workflow, each experiment on a source task in \gevolve{} returns a score that the agent compares with that of a baseline run when planning the next experiment; the adapter computes these scores on the task's held-out targets rather than taking self-reported results, giving up to five scores per source task. On \gtest{} tasks, held-out targets and scores remain invisible, and the agent receives only the outcome of public submission checks.

\noindent{\clubpenalty=10000\textbf{HyperAgents.} HyperAgents \citep{hyperagents2026} maintains an archive of agent implementations and generates candidates by allowing MetaAgent to modify a selected parent's code. TaskAgent revisions change the harness ($H$), while changes to MetaAgent alter the improvement strategy ($I$). Code revisions can also retain reusable experience in repository files ($M$). For evaluation, the TaskAgent is selected from the state with the highest source fitness among the initial agent and valid archived candidates.\par}

We run four candidate generations, with one candidate per generation and tool-call budgets of 400 for TaskAgent and 120 for MetaAgent. The adapter enables native code-execution and editing tools for scientific task solving. As in the native design, the initial agent and each candidate are scored on the held-out targets of every source task in \gevolve{}, giving five scores per source task; MetaAgent reads these per-task metrics and solving traces when generating candidates, and their normalized values serve as archive fitness for parent selection. On \gtest{} tasks, held-out targets and scores remain invisible to TaskAgent.

\FloatBarrier

\section{Variation Across Test Runs}
\label{app:test_variation}

Table~\ref{tab:test_variation} reports the mean task scores in Tables~\ref{tab:modeling_evolution} and~\ref{tab:domain_transfer} with their standard errors over repeated test runs. For task $i$ with $n_i$ valid runs, $\mathrm{Var}(S_i)=s_i^{2}/n_i$, where $s_i^{2}$ is the sample variance of its normalized run scores. For a task with a single valid run, $s_i^{2}$ is the mean sample variance over the tasks of the same agent system and domain. For a mean task score $S$ over $N$ tasks,
\[
\mathrm{SE}(S)=\frac{1}{N}\Big(\sum_{i=1}^{N}\mathrm{Var}(S_i)\Big)^{1/2}.
\]
Tasks without a valid run are excluded, as in the main results.

\begin{table}[!ht]
\centering
\fontsize{7.2}{8.4}\selectfont
\setlength{\tabcolsep}{3pt}
\begin{tabular*}{\textwidth}{@{\extracolsep{\fill}}l*{5}{c}@{}}
\toprule
\textbf{Agent / State} & \multicolumn{5}{c}{\textbf{Mean Task Score $\pm$ SE}} \\
\cmidrule(lr){2-6}
& Overall & Phys. & Life & Earth & Chem. \\
\midrule
\multicolumn{6}{@{}l}{\textbf{EvoScientist}} \\
\quad Initial & 0.5053 $\pm$ 0.0058 & 0.5050 $\pm$ 0.0125 & 0.5001 $\pm$ 0.0027 & 0.5151 $\pm$ 0.0173 & 0.5008 $\pm$ 0.0086 \\
\quad A-evolved & 0.5090 $\pm$ 0.0035 & 0.5266 $\pm$ 0.0097 & 0.4912 $\pm$ 0.0047 & 0.5252 $\pm$ 0.0064 & 0.4928 $\pm$ 0.0063 \\
\quad F-evolved & 0.5112 $\pm$ 0.0040 & 0.5261 $\pm$ 0.0120 & 0.4911 $\pm$ 0.0019 & 0.5159 $\pm$ 0.0080 & 0.5117 $\pm$ 0.0065 \\
\quad Physics-F & 0.5060 $\pm$ 0.0044 & 0.5259 $\pm$ 0.0129 & 0.4889 $\pm$ 0.0061 & 0.5037 $\pm$ 0.0070 & 0.5053 $\pm$ 0.0072 \\
\quad Life-F & 0.5110 $\pm$ 0.0057 & 0.5133 $\pm$ 0.0144 & 0.4891 $\pm$ 0.0036 & 0.5344 $\pm$ 0.0071 & 0.5072 $\pm$ 0.0157 \\
\quad Earth-F & 0.5081 $\pm$ 0.0073 & 0.5344 $\pm$ 0.0166 & 0.4924 $\pm$ 0.0029 & 0.5235 $\pm$ 0.0136 & 0.4819 $\pm$ 0.0195 \\
\quad Chem.-F & 0.5159 $\pm$ 0.0045 & 0.5411 $\pm$ 0.0111 & 0.4936 $\pm$ 0.0041 & 0.5288 $\pm$ 0.0073 & 0.5001 $\pm$ 0.0114 \\
\multicolumn{6}{@{}l}{\textbf{SIA-H}} \\
\quad Initial & 0.5178 $\pm$ 0.0046 & 0.5200 $\pm$ 0.0100 & 0.4987 $\pm$ 0.0085 & 0.5347 $\pm$ 0.0083 & 0.5177 $\pm$ 0.0100 \\
\quad A-evolved & 0.5012 $\pm$ 0.0080 & 0.5135 $\pm$ 0.0236 & 0.4762 $\pm$ 0.0137 & 0.5257 $\pm$ 0.0086 & 0.4920 $\pm$ 0.0173 \\
\quad F-evolved & 0.4995 $\pm$ 0.0094 & 0.4890 $\pm$ 0.0231 & 0.4913 $\pm$ 0.0045 & 0.5249 $\pm$ 0.0231 & 0.4927 $\pm$ 0.0182 \\
\quad Physics-F & 0.5158 $\pm$ 0.0082 & 0.5424 $\pm$ 0.0057 & 0.4984 $\pm$ 0.0090 & 0.5228 $\pm$ 0.0188 & 0.4961 $\pm$ 0.0231 \\
\quad Life-F & 0.5138 $\pm$ 0.0081 & 0.5138 $\pm$ 0.0282 & 0.4962 $\pm$ 0.0042 & 0.5241 $\pm$ 0.0098 & 0.5210 $\pm$ 0.0125 \\
\quad Earth-F & 0.5141 $\pm$ 0.0068 & 0.5220 $\pm$ 0.0204 & 0.5064 $\pm$ 0.0018 & 0.5198 $\pm$ 0.0097 & 0.5081 $\pm$ 0.0149 \\
\quad Chem.-F & 0.5089 $\pm$ 0.0076 & 0.5355 $\pm$ 0.0134 & 0.4860 $\pm$ 0.0162 & 0.5417 $\pm$ 0.0057 & 0.4727 $\pm$ 0.0212 \\
\multicolumn{6}{@{}l}{\textbf{HyperAgents}} \\
\quad Initial & 0.5021 $\pm$ 0.0051 & 0.5080 $\pm$ 0.0119 & 0.4993 $\pm$ 0.0030 & 0.5119 $\pm$ 0.0110 & 0.4892 $\pm$ 0.0123 \\
\quad A-evolved & 0.5037 $\pm$ 0.0041 & 0.5204 $\pm$ 0.0117 & 0.4873 $\pm$ 0.0030 & 0.5262 $\pm$ 0.0064 & 0.4808 $\pm$ 0.0088 \\
\quad F-evolved & 0.4931 $\pm$ 0.0049 & 0.5012 $\pm$ 0.0100 & 0.4841 $\pm$ 0.0034 & 0.5051 $\pm$ 0.0104 & 0.4822 $\pm$ 0.0126 \\
\quad Physics-F & 0.5095 $\pm$ 0.0060 & 0.5334 $\pm$ 0.0077 & 0.4808 $\pm$ 0.0019 & 0.5176 $\pm$ 0.0105 & 0.5063 $\pm$ 0.0201 \\
\quad Life-F & 0.4887 $\pm$ 0.0066 & 0.4612 $\pm$ 0.0214 & 0.4903 $\pm$ 0.0033 & 0.5305 $\pm$ 0.0024 & 0.4726 $\pm$ 0.0151 \\
\quad Earth-F & 0.5003 $\pm$ 0.0069 & 0.5037 $\pm$ 0.0158 & 0.4979 $\pm$ 0.0027 & 0.5157 $\pm$ 0.0089 & 0.4841 $\pm$ 0.0205 \\
\quad Chem.-F & 0.4927 $\pm$ 0.0063 & 0.4897 $\pm$ 0.0146 & 0.4937 $\pm$ 0.0030 & 0.5004 $\pm$ 0.0108 & 0.4871 $\pm$ 0.0171 \\
\multicolumn{6}{@{}l}{\textbf{InternAgent-1.5}} \\
\quad Initial & 0.4903 $\pm$ 0.0084 & 0.4847 $\pm$ 0.0249 & 0.4772 $\pm$ 0.0047 & 0.5047 $\pm$ 0.0181 & 0.4947 $\pm$ 0.0122 \\
\quad A-evolved & 0.4888 $\pm$ 0.0056 & 0.4993 $\pm$ 0.0111 & 0.4805 $\pm$ 0.0019 & 0.4919 $\pm$ 0.0098 & 0.4834 $\pm$ 0.0166 \\
\quad F-evolved & 0.4939 $\pm$ 0.0053 & 0.5026 $\pm$ 0.0158 & 0.4796 $\pm$ 0.0043 & 0.5060 $\pm$ 0.0083 & 0.4873 $\pm$ 0.0104 \\
\quad Physics-F & 0.4935 $\pm$ 0.0053 & 0.5114 $\pm$ 0.0103 & 0.4798 $\pm$ 0.0049 & 0.4953 $\pm$ 0.0134 & 0.4874 $\pm$ 0.0117 \\
\quad Life-F & 0.5001 $\pm$ 0.0051 & 0.5221 $\pm$ 0.0136 & 0.4757 $\pm$ 0.0049 & 0.4978 $\pm$ 0.0099 & 0.5048 $\pm$ 0.0107 \\
\quad Earth-F & 0.4870 $\pm$ 0.0068 & 0.5000 $\pm$ 0.0226 & 0.4827 $\pm$ 0.0049 & 0.5090 $\pm$ 0.0077 & 0.4561 $\pm$ 0.0125 \\
\quad Chem.-F & 0.4943 $\pm$ 0.0080 & 0.5072 $\pm$ 0.0212 & 0.4872 $\pm$ 0.0047 & 0.4794 $\pm$ 0.0230 & 0.5032 $\pm$ 0.0052 \\
\bottomrule
\end{tabular*}
\caption{Mean task scores with standard errors over repeated test runs for the states in Tables~\ref{tab:modeling_evolution} and~\ref{tab:domain_transfer}. Physics-F, Life-F, Earth-F, and Chem.-F denote evolution on the F tasks of one domain.}
\label{tab:test_variation}
\end{table}

\FloatBarrier

\clearpage
\phantomsection
\addcontentsline{toc}{section}{References}
\bibliography{bibliography/custom}
\bibliographystyle{ACM-Reference-Format}

\end{document}